\documentclass[10pt, a4paper]{article}
\usepackage{amsmath}
\usepackage{arydshln}
\usepackage{multirow} 
\usepackage{caption}
\usepackage{booktabs}
\usepackage{graphicx}
\usepackage{enumitem}
\usepackage[table]{xcolor}
\usepackage{colortbl}
\usepackage{multirow}
\usepackage{orcidlink}
\usepackage{array}
\usepackage{booktabs}
\usepackage{hyperref}
\hypersetup{
    colorlinks=false,
    pdfborder={0 0 0}
}

\definecolor{pi}{RGB}{230, 240, 255}   
\definecolor{pp}{RGB}{255, 245, 230}   
\definecolor{nu}{RGB}{240, 240, 240}   

\usepackage{amsmath,amssymb,amsfonts}
\usepackage{textcomp}
\usepackage{xcolor}
\usepackage{tikz}
\usepackage{subcaption}
\usepackage{float}
\usepackage{csquotes}
\usepackage{siunitx}
\usepackage{tabularx}
\usepackage{threeparttable}
\usepackage{multicol}
\usepackage{xcolor,colortbl}
\usepackage{caption}

\usepackage[final]{lrec2026} 

\title{StanceMoE: Mixture-of-Experts Architecture for Stance Detection}

\name{Abdullah Al Shafi, Md. Milon Islam, Sk. Imran Hossain, K. M. Azharul Hasan} 

\address{Department of Computer Science and Engineering, Khulna University of Engineering \& Technology\\
         abdullah@iict.kuet.ac.bd, \{milonislam, imran, az\}@cse.kuet.ac.bd\\}

\abstract{
Actor-level stance detection aims to determine an author’s expressed position toward specific geopolitical actors mentioned or implicated in a text. Although transformer-based models have achieved relatively good performance in stance classification, they typically rely on unified representations that may not sufficiently capture heterogeneous linguistic signals, such as contrastive discourse structures, framing cues, and salient lexical indicators. This motivates the need for adaptive architectures that explicitly model diverse stance-expressive patterns. In this paper, we propose StanceMoE, a context-enhanced Mixture-of-Experts (MoE) architecture built upon a fine-tuned BERT encoder for actor-level stance detection. Our model integrates six expert modules designed to capture complementary linguistic signals, including global semantic orientation, salient lexical cues, clause-level focus, phrase-level patterns, framing indicators, and contrast-driven discourse shifts. A context-aware gating mechanism dynamically weights expert contributions, enabling adaptive routing based on input characteristics. Experiments are conducted on the StanceNakba 2026 Subtask A dataset, comprising 1,401 annotated English texts where the target actor is implicit in the text. StanceMoE achieves a macro-F1 score of 94.26\%, outperforming traditional baselines, and alternative BERT-based variants. 
 \\ \newline  }

\begin{document}
\maketitleabstract
\pagestyle{empty}

\section{Introduction}
Stance detection refers to the automatic identification of an author’s position towards a specific topic, entity, or proposition expressed within a text \cite{garg2024stanceformer}. The task aims to identify whether the presented viewpoint is supportive, opposing, or neutral toward the specific target. The study of stance detection differs from common sentiment analysis because it needs to measure emotional expressions that depend on particular targets, whereas the same text can show different stances based on which target the reader chooses to consider \cite{niu2024challenge}. The nature of this task requires target awareness because it generates both semantic complexity and computational difficulties, particularly when targets remain unmentioned and stances are expressed in an indirect way \cite{kuccuk2020stance}.

To address these challenges, we introduce StanceMoE\footnote{\url{https://github.com/AbdullahRatulk/StanceMoE}}, a context-enhanced Mixture-of-Experts (MoE) architecture designed for actor-level stance detection. Unlike conventional transformer-based approaches that rely on single aggregated representations, our method explicitly decomposes stance modeling into complementary expert modules that capture diverse linguistic and discourse-level phenomena. By incorporating a context-aware gating mechanism, the architecture enables adaptive and input-sensitive fusion of heterogeneous stance signals.
Through comprehensive experimentation and detailed ablation analysis on StanceNakba 2026 Shared Task dataset \cite{stance_nakba_2026}, we demonstrate that explicitly modeling such diverse patterns enables more robust and fine-grained actor-level stance detection with $3^{rd}$ position in the competition.
\vspace{-0.8em}
\section{Related Works}
The field of stance detection has evolved through its transition from rule-based systems \cite{kuccuk2020stance} to classical supervised systems \cite{alturayeif2023systematic}, which rely on manually created features. Deep learning methods, such as Convolutional Neural Networks (CNNs) and Long Short-Term Memory (LSTMs) with attention mechanisms, enable models to automatically extract features and better capture target-specific stance information \cite{gera2025deep}. Current transformer architectures use BERT \cite{garg2024stanceformer} and Large Language Models (LLMs) \cite{pangtey2025large} to deliver excellent results through their ability to understand context and their capacity for few-shot learning.



\begin{figure*}
\centerline{\includegraphics[width=0.92\textwidth]{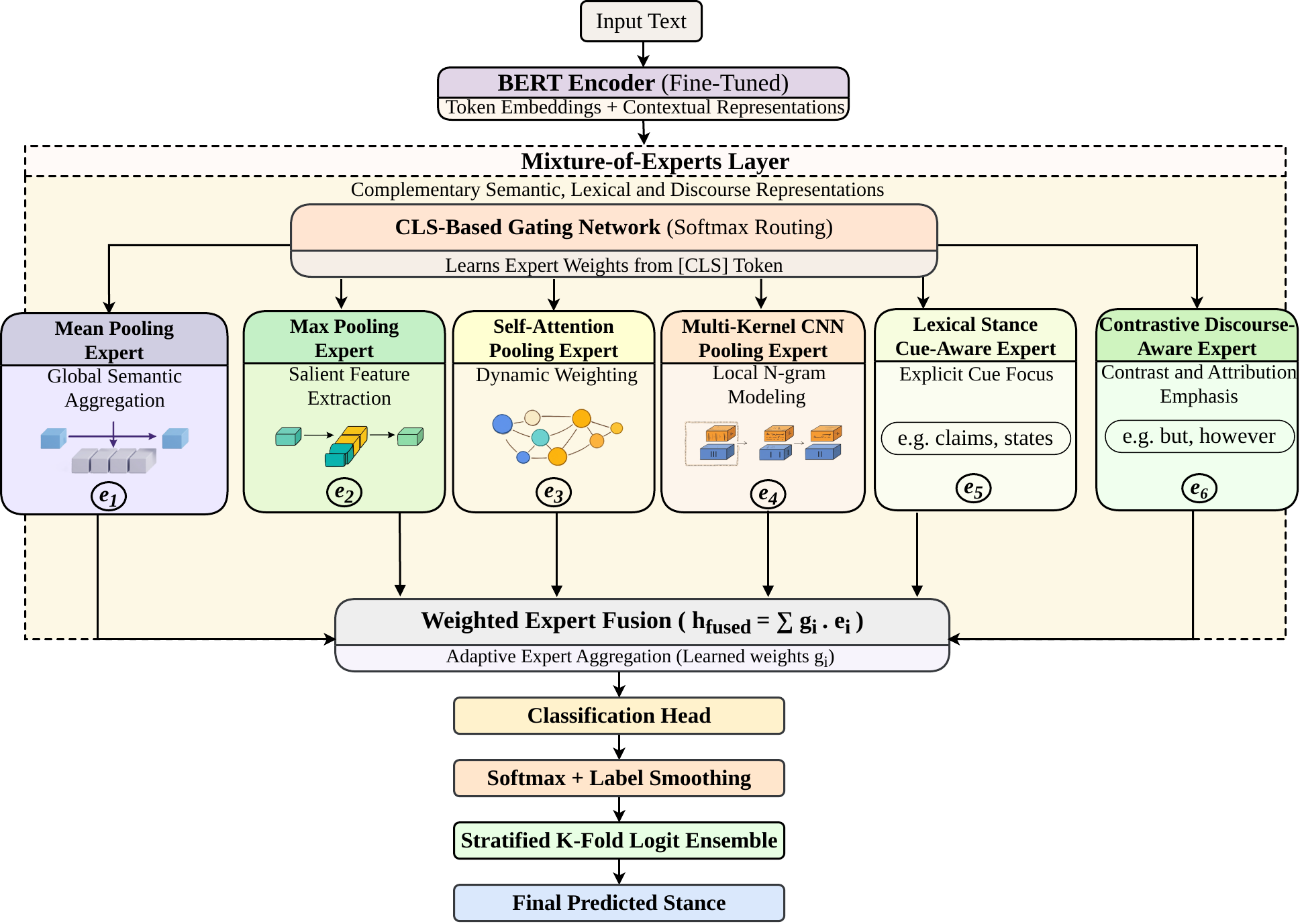}}
\caption{Proposed Mixture-of-Experts architecture for actor-level stance detection.}
\label{fig:moe}
\end{figure*}

\section{Proposed StanceMoE Architecture}
We propose a context-enhanced Mixture-of-Experts architecture developed upon a fine-tuned BERT encoder for stance detection. The overall framework consists of three components: (i) a contextual encoder, (ii) six parallel expert modules capturing complementary linguistic signals, and (iii) a context-aware gating and fusion mechanism. The architecture is illustrated in Fig. \ref{fig:moe}.

\vspace{-1em}
\subsection{Contextual Encoder}
Given an input sequence, we employ a fine-tuned BERT encoder to obtain contextualized token representations and a special \texttt{[CLS]} embedding.

\vspace{-1em}
\subsection{Expert Modules}
While BERT provides promising contextual understanding, stance detection often depends on subtle lexical cues, contrast markers, and discourse patterns. Standard pooling techniques over contextual embeddings may not consistently capture these heterogeneous patterns. To address this, we introduce six complementary experts, each targeting a distinct linguistic phenomenon.

\textbf{Mean Pooling Expert (Global Orientation):} This expert captures the overall semantic direction of a post by averaging token representations. It is effective when stance is consistently expressed throughout the sentence. For example, \textit{“Israel has the right to defend itself.”} (Pro-Israel), or \textit{“Palestinians deserve statehood and equal rights.”} (Pro-Palestine). 

\textbf{Max Pooling Expert (Salient Polarity Tokens):} Some posts contain isolated but highly indicative tokens that determine stance. For instance, the token \textit{“occupation”} serves as the main indicator of stance in the Pro-Palestine statement \textit{“This ongoing occupation must end”}. Similarly, the token \textit{“terrorist”} serves as the main lexical indicator in the Pro-Israel statement \textit{“The terrorist attacks cannot be justified”}. Max pooling selects the most activated feature dimensions to emphasize important tokens throughout the entire sequence.



\textbf{Self-Attention Pooling Expert (Context-Dependent Focus):} The specific clauses of a sentence work as the place where stance exists. The sentence \textit{“While the humanitarian situation is tragic, Israel must respond to security threats”} demonstrate an example of such configurations. In such formulations, one clause presents the main stance while the additional information is presented through the rest of the sentence. The attention mechanism gives more weight to stance components because it assesses their importance throughout the evaluation process.


\textbf{Multi-Kernel CNN Expert (Phrase-Level Patterns):} Geopolitical stance is mostly expressed through short slogan-like phrases which include \textit{“Stand with Israel”} (Pro-Israel) and \textit{“Free Palestine”} (Pro-Palestine). The compact n-grams of the expression serve as effective stance indicators which maintain their function despite minimal contextual information. Multi-kernel convolutions capture such localized phrase patterns.



\textbf{Lexical Cue-Aware Expert (Framing and Reporting Signals):} Neutral posts use reporting language and distancing language to express content like the statement \textit{“According to officials, negotiations are ongoing.”} Verbs like \textit{“claims,” “reports,” “states,”} function as indicators that speakers use indirect methods to present information instead of showing direct support for their statements. This expert combines different ways of showing predetermined cue tokens to improve the ability to separate neutral content from opinionated postings.




\textbf{Contrast-Aware Expert (Discourse Shift Modeling):} Discourse contrast markers such as \textit{“but”} or \textit{“however”} often indicate stance refinement or emphasis. For example, \textit{“I support peace efforts, but the blockade must end.”} or \textit{“Civilian harm is tragic; however, security responses are necessary.”} In such cases, the clause containing the contrast marker often carries stronger stance weight. We therefore amplify representations of contrast tokens to better capture discourse-level stance shifts.

\vspace{-1em}
\subsection{Context-Aware Gating and Fusion}
We introduce a gating mechanism that dynamically weights the experts instead of treating them equally. The final representation is computed as a weighted combination of expert outputs. This design enables adaptive routing. For instance, strongly opinionated posts may rely more on max pooling, contrast-heavy statements may increase the weight of the contrast-aware expert, report-style neutral texts may emphasize cue-aware representations.


The fused representation is then passed to a linear classifier. We train the model using cross-entropy loss with label smoothing. To improve robustness, we employ stratified $K$-fold cross-validation during training and perform weighted logit averaging of those folds during inference. The weighting is done by giving more weight to folds having a higher F1-Score during validation.

\vspace{-1em}
\section{Experimental Setup}
\subsection{Dataset}
The dataset used in this paper is provided by the organizers of the StanceNakba 2026 Shared Task \cite{stance_nakba_2026}. We focus on Subtask A, an actor-level stance detection dataset comprising 1,401 English texts labeled as Pro-Palestine, Pro-Israel, or Neutral. The stance labels denote the author’s alignment with actors, which are implicit in the text and must be inferred by the model. The official split follows a 70/15/15 split for training, development, and test sets. In our experiments, we apply stratified K-fold cross-validation on the combined training and development set (85\% of the data). The official test set (15\%) is kept strictly unseen during training.


\vspace{-1 em}
\subsection{Baseline Methods}
We perform an extensive empirical assessment of existing stance detection techniques, which are classified into three major methodological categories: Machine Learning (ML) methods with TF-IDF features, supervised training with Deep Neural Networks (DNNs), and BERT. The ML methods include Logistic Regression (LR) \cite{rahman2024optimizing}, Multinomial Naive Bayes (MNB) \cite{zannat2025bridging}, Support Vector Machine (SVM) \cite{rahman2024optimizing}, and Random Forest (RF) \cite{shafi2025structured}. We implement several popular DNN architectures as baselines, including Bi-directional Long Short-Term Memory (BiLSTM) \cite{rahman2024optimizing}, Target specific Attentional Network (TAN) \cite{du2017stance},  Gated Convolutional network with Aspect Embedding (GCAE) \cite{xue2018aspect}, Cross network (CrossNet) \cite{du2017stance}. Moreover, we evaluate two alternative variants developed on the BERT backbone: (i) a stacked architecture \cite{khan2025empowering} where expert modules are applied sequentially, and (ii) a feature fusion model \cite{lee2021multimodal} where expert outputs are fused without adaptive weighting.



\begin{table*}[htbp]
\centering
\begin{tabular}{lcccccccc}
\toprule
\multirow{2}{*}{\textbf{Methods}} & \multicolumn{4}{c}{\textbf{K-fold (mean$\pm$std)}} & \multicolumn{4}{c}{\textbf{Weighted Logit Ensemble}}\\
\cmidrule(lr){2-5} \cmidrule(lr){6-9}
& \textit{Acc} & \textit{Pre} & \textit{Rec} & \textit{F1} & \textit{Acc} & \textit{Pre} & \textit{Rec} & \textit{F1} \\
\midrule
LR        & 80.91$\pm$1.79 & 80.90$\pm$1.78 & 80.93$\pm$1.78 & 80.86$\pm$1.78 & 81.28 & 81.25 & 81.33 & 81.22 \\
MNB       & 77.25$\pm$1.64 & 75.88$\pm$2.02 & 78.49$\pm$1.65 & 77.20$\pm$1.65 & 77.38 & 76.02 & 78.63 & 77.33 \\
SVM       & 83.25$\pm$1.71 & 84.73$\pm$1.68 & 83.26$\pm$1.71 & 83.27$\pm$1.72 & 83.37 & 84.94 & 83.35 & 83.43 \\
RF        & 84.05$\pm$1.83 & 84.73$\pm$1.73 & 84.05$\pm$1.82 & 84.05$\pm$1.90 & 84.16 & 84.83 & 84.14 & 84.19 \\
\midrule
BiLSTM    & 85.63$\pm$2.99 & 85.87$\pm$3.00 & 85.63$\pm$2.99 & 85.51$\pm$3.07 & 85.74 & 85.98 & 85.74 & 85.72 \\
TAN      & 85.99$\pm$3.10 & 86.01$\pm$2.81 & 86.36$\pm$2.08 & 85.91$\pm$2.15 & 86.13 & 86.10 & 86.43 & 86.03 \\
GCAE     & 87.69$\pm$2.85 & 87.98$\pm$2.92 & 87.48$\pm$2.16 & 87.49$\pm$2.78 & 87.82 & 88.10 & 87.53 & 87.62 \\
CrossNet  & 85.13$\pm$2.28 & 85.55$\pm$1.89 & 84.90$\pm$1.90 & 84.77$\pm$2.25 & 85.29 & 85.65 & 84.97 & 84.88 \\
\midrule
BERT      & 89.77$\pm$2.35 & 90.04$\pm$2.30 & 89.77$\pm$2.31 & 89.61$\pm$2.29 & 90.05 & 90.28 & 90.03 & 89.86 \\
Stacked   & 91.61$\pm$2.46 & 91.73$\pm$2.43 & 91.62$\pm$2.44 & 91.60$\pm$2.47 & 91.94 & 92.14 & 91.93 & 91.83 \\
Fusion    & 91.03$\pm$2.26 & 91.20$\pm$2.29 & 91.03$\pm$2.24 & 91.02$\pm$2.26 & 91.18 & 91.45 & 91.17 & 91.17 \\
\midrule
\textbf{StanceMoE} & \textbf{94.09$\pm$1.11} & \textbf{94.18$\pm$1.12} & \textbf{94.08$\pm$1.12} & \textbf{94.03$\pm$1.12} & \textbf{94.31} & \textbf{94.45} & \textbf{94.31} & \textbf{94.26} \\
\bottomrule
\end{tabular}
\caption{Performance comparison of baseline models and the proposed StanceMoE on the held-out test set. \textit{“K-fold (mean ± std)”} reports the mean and standard deviation of test performance across fold-specific models, while \textit{“weighted logit ensemble”} is the final ensemble prediction obtained via logit averaging. Here, Acc = Accuracy, Pre = Precision, Rec = Recall, and F1 = F1-Score.}
\label{tab:results}
\end{table*}


\begin{table*}[htbp]
\centering
\begin{tabular}{lccccccccccc}
\toprule
\multirow{2}{*}{\textbf{Methods}} & \multirow{2}{*}{\textit{Acc}} & \multicolumn{3}{c}{\textbf{Pro-Palestine}} & \multicolumn{3}{c}{\textbf{Pro-Israel}} & \multicolumn{3}{c}{\textbf{Neutral}}\\
\cmidrule(lr){3-5} \cmidrule(lr){6-8} \cmidrule(lr){9-11}
& & \textit{Pre} & \textit{Rec} & \textit{F1} & \textit{Pre} & \textit{Rec} & \textit{F1} & \textit{Pre} & \textit{Rec} & \textit{F1} \\
\midrule
w/o Mean & 92.89 & 89.47 & 95.75 & 92.52 & 94.52 & 98.57 & 96.50 & 95.16 & 84.29 & 89.39 \\
w/o Max  & 93.36 & 91.89 & \textbf{95.77} & 93.79 & 92.00 & 98.57 & 95.17 & 96.77 & 85.71 & 90.91 \\
w/o Self-Attention & 91.94 & 89.33 & 94.37 & 91.78 & 93.24 & 98.57 & 95.83 & 93.55 & 82.86 & 87.88 \\
w/o CNN & 93.36 & 92.96 & 92.96 & 92.96 & 93.33 & 99.26 & 96.55 & 93.85 & 87.14 & 90.37 \\
w/o Lexical-cue & 93.36 & 90.54 & 94.37 & 92.41 & \textbf{95.83} & 98.57 & \textbf{97.18} & 95.08 & 82.86 & 88.55 \\
w/o Contrastive & 91.94 & 90.41 & 92.96 & 91.67 & 90.91 & 99.32 & 95.24 & 93.85 & 87.14 & 90.37 \\
\midrule
\textbf{StanceMoE} & \textbf{94.31} & \textbf{94.37} & 94.37 & \textbf{94.37} & 92.11 & \textbf{100} & 95.89 & \textbf{96.88} & \textbf{88.57} & \textbf{92.54} \\
\bottomrule
\end{tabular}
\caption{Classwise ablation study of StanceMoE using weighted logit ensemble, showing the effect of removing individual expert modules.}
\label{tab:ablation}
\end{table*}

\vspace{-0.9 em}
\section{Experimental Results}
\subsection{Actor-level Stance Detection}
Table \ref{tab:results} presents the comparative performance of all models for actor-level stance detection. Among traditional baselines, RF achieves the best results with an F1-Score of 84.19\%, which shows that ensemble tree-based methods perform better than other methods. Among neural models, GCAE performs highest (87.62\% F1), which shows that attention-based architectures perform well to model stance. The use of contextual embeddings results in better performance outcomes. BERT achieves an F1-Score of 89.86\%, while expert-based architectures like Stacked (91.83\%) and Fusion (91.17\%) show promising results, but their improvements remain within fixed expert integration boundaries.

The proposed StanceMoE achieves the best performance with an F1-Score of 94.26\%, which surpasses all other cases of comparison. Additionally, our team has stood $3^{rd}$ in the competition. The system demonstrates performance enhancement because of its adaptive expert weighting system, which uses a gating mechanism. The proposed system also shows stable performance with low cross-validation variance ($\approx \pm 1.1$\%).

\subsection{Ablation Study}
Table \ref{tab:ablation} presents an ablation analysis to examine the contribution of each expert in the proposed StanceMoE architecture. Removing any expert generally degrades overall performance, confirming that the experts capture complementary stance signals. The largest drops occur when removing the self-attention or contrastive experts, indicating their importance in modeling contextual dependencies and discourse-level stance shifts. Excluding the mean or max pooling experts also reduces performance, suggesting the importance of salient lexical cues and global contextual signals for stance identification.

The lexical-cue expert particularly benefits the Pro-Palestine class, while the contrastive expert improves recognition of stance transitions and argumentative structures. Although some variants maintain competitive scores for individual classes, they generally exhibit reduced balance across classes, especially for the neutral category. Overall, the full StanceMoE model achieves the best and most balanced performance across all classes, demonstrating the effectiveness of adaptive expert integration.

\section{Conclusion}
In this paper, we introduced StanceMoE, a context-enhanced MoE architecture for stance detection. Motivated by the limitations of unified transformer representations capturing heterogeneous linguistic phenomena, our approach explicitly models complementary stance-indicative signals through expert modules and integrates them via a context-aware gating mechanism. This design enables adaptive modeling of lexical, semantic, and discourse-level patterns that characterize stance expression in sensitive contexts. Extensive experiments demonstrate that StanceMoE substantially outperforms traditional baseline models. Ablation analysis further confirms the complementary contributions of individual experts and the importance of adaptive expert fusion for robust detection. Overall findings highlight the effectiveness of explicitly disentangling and dynamically integrating diverse linguistic signals for stance detection.

\section*{Bibliographical References}
\label{sec:reference}
\vspace{-2.7em}
\bibliographystyle{lrec2026-natbib}
\bibliography{lrec2026-example_nakba}

@inproceedings{stance_nakba_2026,
     author    = {Aldous, Kholoud Khalil and Biswas, Md Rafiul and Bessghaier, Mabrouka and Ibrahim, Shimaa and Attia, Kais and Zaghouani, Wajdi},
     title     = {{StanceNakba} Shared Task: Actor and Topic-Aware Stance Detection in Public Discourse},
     booktitle = {Proceedings of the 15th International Conference on Language Resources and Evaluation (LREC'26)},
     month     = May,
     year      = {2026},
     address   = {Palma, Spain}
   }

@article{alturayeif2023systematic,
  title={A systematic review of machine learning techniques for stance detection and its applications},
  author={Alturayeif, Nora and Luqman, Hamzah and Ahmed, Moataz},
  journal={Neural Computing and Applications},
  volume={35},
  number={7},
  pages={5113--5144},
  year={2023},
  publisher={Springer}
}

@article{kuccuk2020stance,
  title={Stance detection: A survey},
  author={K{\"u}{\c{c}}{\"u}k, Dilek and Can, Fazli},
  journal={ACM Computing Surveys},
  volume={53},
  number={1},
  pages={1--37},
  year={2020},
  publisher={ACM New York, NY, USA}
}

@article{gera2025deep,
  title={Deep Learning in Stance Detection: A Survey},
  author={Gera, Parush and Neal, Tempestt},
  journal={ACM Computing Surveys},
  volume={58},
  number={1},
  pages={1--37},
  year={2025},
  publisher={ACM New York, NY}
}

@article{burnham2025stance,
  title={Stance detection: a practical guide to classifying political beliefs in text},
  author={Burnham, Michael},
  journal={Political Science Research and Methods},
  volume={13},
  number={3},
  pages={611--628},
  year={2025},
  publisher={Cambridge University Press}
}

@article{pangtey2025large,
  title={Large language models meet stance detection: A survey of tasks, methods, applications, challenges and future directions},
  author={Pangtey, Lata and Bhatnagar, Anukriti and Bansal, Shubhi and Dar, Shahid Shafi and Kumar, Nagendra},
  journal={arXiv:2505.08464},
  year={2025}
}

@inproceedings{garg2024stanceformer,
  title={Stanceformer: Target-aware transformer for stance detection},
  author={Garg, Krishna and Caragea, Cornelia},
  booktitle={Findings of the Association for Computational Linguistics: EMNLP 2024},
  pages={4969--4984},
  year={2024}
}

@inproceedings{niu2024challenge,
  title={A challenge dataset and effective models for conversational stance detection},
  author={Niu, Fuqiang and Yang, Min and Li, Ang and Zhang, Baoquan and Peng, Xiaojiang and Zhang, Bowen},
  booktitle={Proceedings of the 2024 Joint International Conference on Computational Linguistics, Language Resources and Evaluation (LREC-COLING 2024)},
  pages={122--132},
  year={2024}
}

@article{khan2025empowering,
  title={Empowering Urdu sentiment analysis: an attention-based stacked CNN-Bi-LSTM DNN with multilingual BERT},
  author={Khan, Lal and Qazi, Atika and Chang, Hsien-Tsung and Alhajlah, Mousa and Mahmood, Awais},
  journal={Complex \& Intelligent Systems},
  volume={11},
  number={1},
  pages={10},
  year={2025},
  publisher={Springer}
}

@inproceedings{rahman2024optimizing,
  title={Optimizing SMS spam detection: comparative analysis of hybrid voting ensembles and bi-LSTM networks with stratified cross-validation},
  author={Rahman, Arifur and Parvej, Shahriar and Alam, Kazi Saeed and Fattah, HM Abdul},
  booktitle={2024 5th International Conference on Data Intelligence and Cognitive Informatics (ICDICI)},
  pages={1030--1035},
  year={2024},
  organization={IEEE}
}

@article{lee2021multimodal,
  title={Multimodal emotion recognition fusion analysis adapting BERT with heterogeneous feature unification},
  author={Lee, Sanghyun and Han, David K and Ko, Hanseok},
  journal={IEEE Access},
  volume={9},
  pages={94557--94572},
  year={2021},
  publisher={IEEE}
}

@article{ma2025exploring,
  title={Exploring multi-agent debate for zero-shot stance detection: A novel approach},
  author={Ma, Junxia and Wang, Changjiang and Rong, Lu and Wang, Bo and Xu, Yaoli},
  journal={Applied Sciences},
  volume={15},
  number={9},
  pages={4612},
  year={2025},
  publisher={MDPI}
}

@article{zhang2024survey,
  title={A survey of stance detection on social media: New directions and perspectives},
  author={Zhang, Bowen and Dai, Genan and Niu, Fuqiang and Yin, Nan and Fan, Xiaomao and Wang, Senzhang and Cao, Xiaochun and Huang, Hu},
  journal={arXiv:2409.15690},
  year={2024}
}

@article{yang2025llm,
  title={LLM-enhanced multiple instance learning for joint rumor and stance detection with social context information},
  author={Yang, Ruichao and Ma, Jing and Gao, Wei and Lin, Hongzhan},
  journal={ACM Transactions on Intelligent Systems and Technology},
  volume={16},
  number={3},
  pages={1--27},
  year={2025},
  publisher={ACM New York, NY}
}

@article{guo2022survey,
  title={A survey on automated fact-checking},
  author={Guo, Zhijiang and Schlichtkrull, Michael and Vlachos, Andreas},
  journal={Transactions of the Association for Computational Linguistics},
  volume={10},
  pages={178--206},
  year={2022}
}

@article{alsmadi2024stance,
  title={Stance detection in the context of fake news—A new approach},
  author={Alsmadi, Izzat and Alazzam, Iyad and Al-Ramahi, Mohammad and Zarour, Mohammad},
  journal={Future Internet},
  volume={16},
  number={10},
  pages={364},
  year={2024},
  publisher={MDPI}
}

@inproceedings{zannat2025bridging,
  title={Bridging the gap in bangla healthcare: Machine learning based disease prediction using a symptoms-disease dataset},
  author={Zannat, Rowzatul and Al Shafi, Abdullah and Muntakim, Abdul},
  booktitle={2025 International Conference on Electrical, Computer and Communication Engineering (ECCE)},
  pages={1--6},
  year={2025},
  organization={IEEE}
}

@inproceedings{zhu2025ratsd,
  title={RATSD: Retrieval augmented truthfulness stance detection from social media posts toward factual claims},
  author={Zhu, Zhengyuan and Zhang, Zeyu and Zhang, Haiqi and Li, Chengkai},
  booktitle={Findings of the Association for Computational Linguistics: NAACL 2025},
  pages={3366--3381},
  year={2025}
}

@article{shafi2025structured,
  title={A structured dataset of disease-symptom associations to improve diagnostic accuracy},
  author={Shafi, Abdullah Al and Zannat, Rowzatul and Muntakim, Abdul and Hasan, Mahmudul},
  journal={arXiv:2506.13610},
  year={2025}
}

@inproceedings{du2017stance,
  title={Stance classification with target-specific neural attention networks},
  author={Du, Jiachen and Xu, Ruifeng and He, Yulan and Gui, Lin},
  booktitle={2017 26th International Joint Conference on Artificial Intelligence (IJCAI)},
  pages={3988--3994},
  year={2017}
}

@inproceedings{xue2018aspect,
  title={Aspect based sentiment analysis with gated convolutional networks},
  author={Xue, Wei and Li, Tao},
  booktitle={Proceedings of the 56th Annual Meeting of the Association for Computational Linguistics (Volume 1: Long Papers)},
  pages={2514--2523},
  year={2018}
}

\appendix
\vspace{-1em}
\section{Need of Stance Detection}
The ability to detect stance has become increasingly important due to the growth of online discourse. Applications span multiple domains, including electoral trend analysis \cite{burnham2025stance}, examination of argumentative interactions in online debates \cite{ma2025exploring}, monitoring of social media discussions \cite{zhang2024survey}, rumor assessment and verification \cite{yang2025llm}, automated fact-checking systems \cite{guo2022survey}, fake news identification \cite{alsmadi2024stance}, and stance-aware information retrieval \cite{zhu2025ratsd}. In these contexts, understanding directional opinion toward a claim or actor is more informative than simply measuring sentiment \cite{garg2024stanceformer}.



\section{Operational Details of Expert Modules}
Given an input sequence $X = \{x_1, \dots, x_T\}$, pre-trained BERT encoder obtains contextualized token representations: $H = \{h_1, h_2, \dots, h_T\}, \quad h_i \in \mathbb{R}^d$
where $d$ denotes the hidden dimension and $T$ is the length of the tokenized input sequence.

\textbf{Mean Pooling Expert:} This expert captures global semantic information as shown in \eqref{eq:mean}.
\begin{equation}
    e_1 = W_1 \left( \frac{1}{T} \sum_{i=1}^{T} h_i \right)
    \label{eq:mean}
\end{equation}

\textbf{Max Pooling Expert:} To extract salient token-level features, we utilize max pooling expert as in \eqref{eq:max}.
\begin{equation}
    e_2 = W_2 \left( \max_{i=1}^{T} h_i \right)
    \label{eq:max}
\end{equation}

\textbf{Self-Attention Pooling Expert:} We introduce a trainable attention vector $v$ to compute token importance as mentioned in \eqref{eq:sa}.
\begin{equation}
    \alpha_i = \frac{\exp(\tanh(h_i^\top v))}{\sum_{j=1}^{T} \exp(\tanh(h_j^\top v))}
    \label{eq:sa}
\end{equation}
\begin{equation}
    e_3 = W_3 \left( \sum_{i=1}^{T} \alpha_i h_i \right)
\end{equation}

\textbf{Multi-Kernel CNN Expert:} To model local n-gram patterns, we apply one-dimensional convolutions with kernel sizes $k \in \{2,3,4,5\}$ as demonstrated in \eqref{eq:cnn}.
\begin{equation}
    e_4 = W_4 \left( \text{Concat}\big( \text{MeanPool}(\text{ReLU}(\text{Conv}_k(H))) \big) \right)
    \label{eq:cnn}
\end{equation}

\textbf{Lexical Cue-Aware Expert:} We define a set of stance-indicative lexical cues and generate a binary mask over their positions $C$. The cue-aware representation is computed as \eqref{eq:cue}.
\begin{equation}
    e_5 = W_5 \left( \frac{\sum_{i \in C} h_i}{|C| + \epsilon} \right)
    \label{eq:cue}
\end{equation}
where, $\epsilon$ avoids division by zero.

\textbf{Contrast-Aware Expert:} To model discourse contrast markers (e.g., ``but'', ``however''), we enhance their contextual influence. Let, $D$ denote the set of contrast token positions. The overall process is shown in \eqref{eq:contrast}.
\begin{equation}
    \tilde{h}_i =
    \begin{cases}
    3h_i & \text{if } i \in D, \\
    h_i & \text{otherwise},
    \end{cases}
\end{equation}
\begin{equation}
    e_6 = W_6 \left( \frac{\sum_{i=1}^{T} \tilde{h}_i}{|D| + \epsilon} \right)
    \label{eq:contrast}
\end{equation}

\section{Operational Details of Context-Aware Gating}
The gating network takes the `[CLS]` representation $h_{\text{cls}} \in \mathbb{R}^d$ from the BERT encoder as input. 
A learnable linear layer $W_g \in \mathbb{R}^{N \times d}$ with bias $b_g \in \mathbb{R}^{N}$ computes logits for $N=6$ experts, which are normalized via a softmax to generate the gating weights as in \eqref{eq:gating1}.
\begin{equation}
    g = \text{Softmax}(W_g h_{\text{cls}} + b_g)
    \label{eq:gating1}
\end{equation}
where $g \in \mathbb{R}^{K}$, $g_i \ge 0$ for all $i$, and $\sum_{i=1}^{K} g_i = 1$. 
Each gating weight $g_i$ determines the contribution of the corresponding expert output $e_i$ to the final representation, which is obtained via weighted aggregation using \eqref{eq:moe}.
\begin{equation}
    h_{\text{moe}} = \sum_{i=1}^{K} g_i \, e_i
    \label{eq:moe}
\end{equation}
The fused representation $h_{\text{moe}}$ is then passed through a task-specific linear layer with learnable weights $W_o$ and bias $b_o$ to generate the model predictions through \eqref{eq:y}.
\begin{equation}
    \hat{y} = \text{softmax}(W_o h_{\text{moe}} + b_o)
    \label{eq:y}
\end{equation}
where $\hat{y}$ represents the predicted class probabilities. The parameters $W_g$, $b_g$, $W_o$, and $b_o$ are all learnable and optimized jointly with the rest of the network via backpropagation, allowing the model to learn both how to combine experts and how to make accurate predictions.

\section{Hyperparameters}
Table \ref{tab:hyperparameters} shows the hyperparameters used in the experiment.

\begin{table}[h]
\centering
\begin{tabular}{lc}
\toprule
\textbf{Hyperparameter} & \textbf{Value} \\
\midrule
Max sequence length & 128 \\
Batch size & 16 \\
Number of epochs & 10 \\
Learning rate & $5\times10^{-5}$ \\
Number of splits ($k$) & 10 \\
Label smoothing factor & 0.25 \\
Random seed & 42 \\
\bottomrule
\end{tabular}
\caption{Training hyperparameters used in StanceMoE.}
\label{tab:hyperparameters}
\end{table}

\begin{figure}
\centerline{\includegraphics[width=0.46\textwidth]{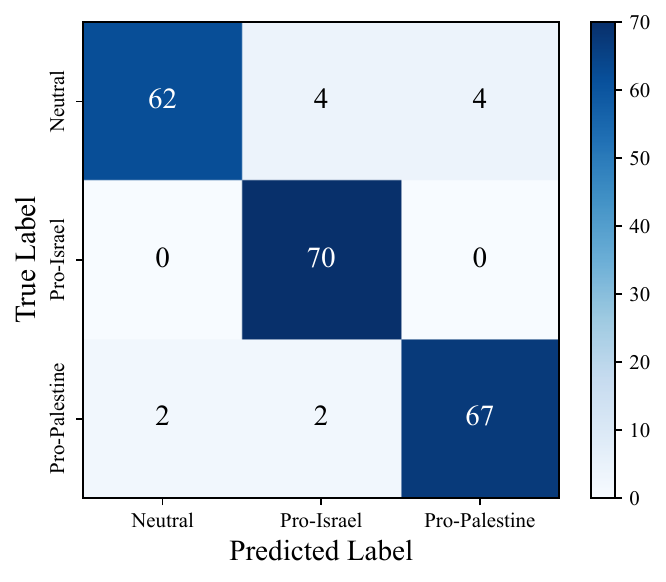}}
\caption{Confusion matrix using the proposed StanceMoE architecture.}
\label{fig:cm}
\end{figure}

\begin{table*}[htbp]
\centering
\resizebox{\textwidth}{!}{%
\begin{tabular}{lcccccccc}
\toprule
\multirow{2}{*}{\textbf{Methods}} & \multicolumn{4}{c}{\textbf{K-fold (mean$\pm$std)}} & \multicolumn{4}{c}{\textbf{Weighted Logit Ensemble}}\\
\cmidrule(lr){2-5} \cmidrule(lr){6-9}
& \textit{Acc} & \textit{Pre} & \textit{Rec} & \textit{F1} & \textit{Acc} & \textit{Pre} & \textit{Rec} & \textit{F1} \\
\midrule
w/o Mean & 91.75$\pm$1.76 & 92.02$\pm$1.54 & 91.74$\pm$1.77 & 91.65$\pm$1.86 & 92.89 & 93.05 & 92.88 & 92.80 \\
w/o Max  & 91.52$\pm$1.23 & 91.83$\pm$1.06 & 91.51$\pm$1.23 & 91.43$\pm$1.30 & 93.36 & 93.56 & 93.35 & 93.29 \\
w/o Self-att & 91.47$\pm$1.04 & 91.65$\pm$0.99 & 91.46$\pm$1.04 & 91.38$\pm$1.08 & 91.94 & 92.04 & 91.93 & 91.83 \\
w/o CNN & 92.09$\pm$1.04 & 92.20$\pm$0.97 & 92.08$\pm$1.04 & 92.00$\pm$1.11 & 93.36 & 93.38 & 93.37 & 93.29 \\
w/o Lexical-cue & 91.70$\pm$1.84 & 91.83$\pm$1.84 & 91.70$\pm$1.84 & 91.67$\pm$1.83 & 93.36 & 93.41 & 93.36 & 93.32 \\
w/o Contrastive & 91.75$\pm$1.81 & 91.99$\pm$1.67 & 91.74$\pm$1.82 & 91.64$\pm$1.87 & 91.94 & 92.13 & 91.94 & 91.82 \\
\midrule
\textbf{StanceMoE} & \textbf{94.09$\pm$1.11} & \textbf{94.18$\pm$1.12} & \textbf{94.08$\pm$1.12} & \textbf{94.03$\pm$1.12} & \textbf{94.31} & \textbf{94.45} & \textbf{94.31} & \textbf{94.26} \\
\bottomrule
\end{tabular}%
}
\caption{Overall K-fold and weighted ensemble ablation study on the test set.}
\label{tab:detailedablation}
\end{table*}


\begin{table*}[t]
\centering
\small
\begin{tabular}{p{7.5cm} p{1.4cm} p{1.4cm} p{3.8cm}}
\hline
\textbf{Text} & \textbf{Actual} & \textbf{Predicted} & \textbf{Primary Error Source} \\
\hline
This is shameful. And the Jewish people, God's chosen people. May God bless and protect you. 
& \cellcolor{nu}{Neutral} & \cellcolor{pi}{Pro-Israel} 
& Religious praise misinterpreted as political stance due to strong positive lexical cues. \\
\hline
Most comments are clothed with sycophancy. Nigerians have a right of association, but you can stand with Gaza. The people of Gaza started this aggressive behaviour on 7th October when they launched a coordinated attack against the state of Israel. 
& \cellcolor{pp}Pro-Palestine & \cellcolor{nu}Neutral 
& Complex multi-clause reasoning; insufficient emphasis on dominant stance clause. \\
\hline
If you are talking about the doctor not talking about Palestine. He is German living in Germany where any anti-Jewish sentiment is taken very seriously and to many people anti-Israel equals antisemitism, so it is in his better interest not to speak about it at all, not necessarily meaning he supports anyone. 
& \cellcolor{nu}Neutral & \cellcolor{pi}Pro-Israel 
& Confusion between topic discussion and stance expression due to keyword activation. \\
\hline
Against the Islamist terrorists who hate Jews and the West. 
& \cellcolor{nu}Neutral & \cellcolor{pp}Pro-Palestine 
& Short ambiguous input; polarity incorrectly associated with a stance. \\
\hline
As I stood up with BLM, I stand against oppression in all forms. Release Palestinian hostages before talking about anything else. 
& \cellcolor{nu}Neutral & \cellcolor{pp}Pro-Palestine 
& Humanitarian framing interpreted as explicit alignment due to lexical cues. \\
\hline
I was raised Muslim, and I know intimately how oppressive the religion can be in various ways. That being said, because I understand that Palestinian gay people are being oppressed for more than their sexuality right now, and that matters too. 
& \cellcolor{pp}Pro-Palestine & \cellcolor{nu}Neutral 
& Failure to capture nuanced contrastive stance; insufficient weighting of discourse signals. \\
\hline
\end{tabular}
\caption{Representative genuine misclassification cases of StanceMoE. Different background colors are used for visual distinction.}
\label{error}
\end{table*}

\vspace{-2em}
\section{Baseline ML methods}
We utilize four popular ML models as our baseline systems: (1) Logistic Regression (LR): A linear model that predicts class probabilities using weighted features, effective for high-dimensional text data. (2) Multinomial Naive Bayes (MNB): A probabilistic model based on word frequency distributions, assuming feature independence, well-suited for text classification. (3) Support Vector Machine (SVM): A margin-based classifier that finds the optimal hyperplane to separate classes, robust in sparse text feature spaces. (4) Random Forest (RF): An ensemble of decision trees that improves classification performance by reducing overfitting through bagging and feature randomness.

\section{Baseline DNN}
We use four popular deep neural network models as our baseline systems: (1) BiLSTM \cite{rahman2024optimizing} is used to create bi-directional context models which predict stances without requiring to use target information. (2) GCAE \cite{xue2018aspect} provides a convolutional system which employs a gating mechanism to block non-target features. (3) TAN \cite{du2017stance} uses an attention-boosted BiLSTM to identify important contextual details which help to determine stances. (4) CrossNet \cite{du2017stance} enhances attention processing through an aspect-based attention layer which operates before the classification process to enhance target-based feature extraction.


\section{Evaluation Metrics}
We evaluate model performance using macro-F1 score as the primary metric. Additionally, accuracy, macro precision, and macro recall are reported.

\section{Overall Ablation Study}
Table \ref{tab:detailedablation} shows the overall ablation study on the test set, rather than class-specific one.

\section{Confusion matrix}
Fig. \ref{fig:cm} shows the confusion matrix using proposed StanceMoE architecture.

\section{Error analysis}
The qualitative error analysis in Table \ref{error} reveals several systematic patterns.

(1) The model occasionally exhibits an excessive dependence on strong lexical polarity cues. The first and fifth examples show religious praise as well as humanitarian language which are mistakenly interpreted as explicit political alignment. 



(2) Meta-discursive commentary is sometimes confused with stance. The text in the third example describes antisemitism together with social norms without showing any specific support. The system uses target-related keywords to make stance predictions about the content. This is difficult because it requires the model to differentiate between entity mentioned and evaluative positioning.


(3) Complex multi-clause and contrastive reasoning remains challenging. The second and sixth examples contain intricate discourse structures that display stance through their multiple levels of argumentation. The system design includes experts to handle both attention and contrast decoding tasks, but in some cases the gating system fails to deliver proper evaluation for these specialized discourse elements.


(4) Very short and context-poor statements like the fourth example are inherently ambiguous. In such cases, the model appears to associate the polarity with dominant training patterns rather than grounding the prediction in the explicit evaluation of the target.

Overall, these errors suggest that while StanceMoE is effective for explicit and lexically polarized stance expressions, nuanced and implicitly framed discourse remains a challenging direction for future improvement.

\vspace{-1.5em}
\section{Future Work}
Future research may extend this framework toward explicitly target-aware stance modeling or explore cross-topic and cross-domain stance detection to further evaluate the robustness and generalization capacity of the proposed architecture. Investigating multilingual adaptation may also enhance its applicability to broader geopolitical discourse contexts.

\vspace{-1em}
\section{Acknowledgement}
We would like to express our sincere gratitude to the organizer of StanceNakba 2026 shared task, and anonymous reviewers for their support. The authors acknowledge the use of ChatGPT (OpenAI) as an assistive tool for language refinement, code suggestions, and conceptual structuring. All generated content was critically reviewed, validated, and appropriately adapted by the authors, who take full responsibility for the accuracy and integrity of the work.


\end{document}